\documentclass[manuscript]{elsarticle}

\usepackage{lineno,hyperref}

\usepackage{amsmath,amssymb, subcaption}

\journal{Journal of Biomedical Informatics}

\begin{document}

\begin{frontmatter}

\title{Justifying Diagnosis Decisions by Deep Neural Networks}

\author{Graham Spinks\fnref{myfootnote}}
\fntext[myfootnote]{graham.spinks@cs.kuleuven.be}

\author{Marie-Francine Moens\fnref{myfootnote2}}
\fntext[myfootnote2]{sien.moens@cs.kuleuven.be}
\address{Department of Computer Science - LIIR, KU Leuven}

\begin{abstract}
An integrated approach is proposed across visual and textual data to both determine and justify a medical diagnosis by a neural network. As deep learning techniques improve, interest grows to apply them in medical applications. To enable a transition to workflows in a medical context that are aided by machine learning, the need exists for such algorithms to help justify the obtained outcome so human clinicians can judge their validity. In this work, deep learning methods are used to map a frontal X-Ray image to a continuous textual representation. This textual representation is decoded into a diagnosis and the associated textual justification that will help a clinician evaluate the outcome. Additionally, more explanatory data is provided for the diagnosis by generating a realistic X-Ray that belongs to the nearest alternative diagnosis. With a clinical expert opinion study on a subset of the X-Ray data set from the Indiana University hospital network, we demonstrate that our justification mechanism significantly outperforms existing methods that use saliency maps. While performing multi-task training with multiple loss functions, our method achieves excellent diagnosis accuracy and captioning quality when compared to current state-of-the-art single-task methods. 
\end{abstract}


\end{frontmatter}


\section{Introduction}

In this paper, we suggest a justification technique that provides both textual and visual evidence for the diagnosis made by a neural network on a set of thorax X-Rays. In recent years interest has grown in applying machine learning techniques to clinical problems where data is often complex and unstructured \citep{miotto2017deep}. As volumes of medical data are expected to increase, there are multiple benefits to the improvement of machine learning tools for diagnosis and report generation. In high volume applications, a semi-autonomous workflow could lead to big economic and time savings. Additionally, a better informed decision process should lead to better health outcomes for patients. 

While there have been many recent attempts to classify X-Rays according to diagnosis with (deep) neural networks (\citep{shin2016learning}, \citep{islam2017abnormality}, \citep{salehinejad2017interpretation}) current solutions often have limitations or are difficult to interpret. The amount of labeled data in such applications is frequently limited and multiple labels per sample are often present which complicates the diagnosis task. These issues make it difficult to apply such techniques in real-world situations, especially in medical applications where wrong decisions might have large impacts. 

This has led to the search of justification techniques that present evidence that help explain the outcome. For visual inputs, this usually consists of heatmaps, while for text, the portion of the text that was most relevant can be highlighted. While these methods provide support for clinicians, novel justification techniques remain of large interest. This research focuses on implementing an artificial neural network that can aid clinicians and ease their workload in X-Ray diagnosis. To help understand the reached conclusion of the network, both a textual and visual justification technique are included in this approach. 

Due to the different nature of text and image, they usually require different machine learning solutions. While a sentence contains a sequence of discrete symbols from a large vocabulary, images contain adjacent pixels with continuous color values. Our approach attempts to bridge both modalities by learning a continuous vector representation for a sentence that serves as an intermediate representation between the discrete textual symbols and the continuous image pixels. To create such representations, which we will refer to as `embeddings', we use Adversarially Regularized Autoencoders (ARAE) \citep{junbo2017adversarially} which have the benefit that similar sentences are mapped to a similar location in the continuous representation space. 

On completion of training, the system is thus able to map an X-Ray to a corresponding continuous vector that can be decoded into the suitable diagnosis as well as a caption that functions as a primary justification of why the network reached a particular conclusion. 

Additionally, the system learns to generate new and realistic X-Ray images given a certain diagnosis and label. Thus, we can visually present evidence for what the network expects an X-Ray to look like for the nearest alternative diagnosis. The generated image is a realistic yet unseen X-Ray generated from an adversarial training setup \citep{2014_GAN_original_paper}. 

Due to the holistic nature of our entire setup, a human operator wishing to evaluate an X-Ray image would thus retain the following evidence: 

\begin{itemize}
	\item A diagnosis label inferred from the visual input;
	\item A textual caption that explains which visual information was found in the input image;
	\item An X-Ray image that is very similar to the input image but belongs to the nearest alternative diagnosis.
\end{itemize}

The contribution of this work is the elaboration of a holistic system that performs classification (diagnosis) as well as generation across visual and textual domains. Our system includes complementary textual and visual justification of the obtained diagnosis, allowing a user to quickly and accurately understand and analyze why the network chooses one outcome over an other and additionally judge the correctness. In an evaluation with human experts, we show that our novel justification approach significantly outperforms existing methods using saliency maps. We also show that our approach, that achieves superior justification outcomes by performing multi-task training with multiple components, achieves excellent performance when compared to recent single-task baselines in classification and image captioning tasks.

\section{Motivation and Background}

Diagnosing in the medical field is an application where a degree of automation could make a huge difference, due both to the high volume of such tasks that require highly trained clinicians, as well as the high cost of faulty classifications \citep{lee2017deep}. An issue for machine learning algorithms is that they rely on annotated data sets which are often not perfect and might perpetrate errors. Previous works have claimed to outperform traditional radiologists when classifying X-Rays \citep{rajpurkar2017chexnet}, yet some researchers have pointed out that the labels in the underlying data set don't always line up correctly with the images \citep{oakdenrayner2018chexnet} leading to doubts about the significance of the results. The cause might be simple errors in annotation or something more complex, such as additional sources of evidence that weren't incorporated in the data set. Algorithms that rely on such annotations thus propagate some of the underlying issues and therefore require a human operator to verify the outcome. All of these issues point to the complexities of diagnosing medical data sets and the need for improved methodologies to explain or justify the outcomes to clinicians.  

This research will focus on a methodology for diagnosis and justification in a setting where such issues are challenging. More specifally, we use the data set of the National Library Medicine, National Institutes of Health, Bethesda, MD, USA \citep{demner2015preparing} which contains 3955 radiology reports and 7470 associated chest X-Rays, a relatively limited amount of data compared to typical deep learning data sets.  Despite some concerns surrounding the annotated labels \citep{oakdenrayner2017exploring}, such a setting is in fact favorable to study a justification technique such as ours. If an algorithm leads to a particular conclusion where the ground truth label is inaccurate, a clear justification mechanism can help clarify such issues. Note that other researchers have worked on this data set before, such as \citet{shin2016learning}, who train a network to annotate the data set with multiple diagnosis labels. \citet{islam2017abnormality} attempt to classify the labels using deep learning methods and \citet{wang2017chestx} use this data set as a gold standard to test their diagnosis results for a network that is trained on a larger X-Ray data set.

To provide some background to the methods of this paper, we will first discuss some existing approaches to medical image diagnosis and justification. 

\subsection{Medical Diagnosis and Justification}

To obtain a good classification result in itself, most approaches in recent years have moved on from more traditional machine learning approaches as Naive Bayes and Support Vector Machines to incorporate Artificial Neural Networks (ANN). A special subset of those ANNs, Convolutional Neural Networks (CNN) have been particularly successful when dealing with visual input. Another subset of ANNs, Recurrent Neural Networks (RNN) have been shown to be particularly adept at generating or analyzing sequences, such as discrete texts. An often used type of RNN is the Long Short-Term Memory network (LSTM) \citep{gers2000learning}. The aforementioned methods have been widely applied in the medical domain to tasks such as image segmentation, computer-aided detection, diagnosis, labeling and captioning \citep{lee2017deep}.

Some approaches to image diagnosis attempt to map images directly to a label \citep{wang2017chestx}. An issue with most medical data sets is that several disease markers might be present in one sample, which means multiple labels need to be found per image which directly complicates the task. While \citet{wang2017chestx} approach this problem with a multi-label classification loss, others resolve this by creating medical reports or annotations directly from the image, from which several diagnoses might be derived \citep{shin2016learning, jing2017automatic, zhang2017mdnet}. 

With regard to justification of an outcome, most researchers have focused on overlaying some type of attention- or heatmap, also called saliency map, to indicate the main contributing factors for the outcome \citep{tataru2017deep, islam2017abnormality, wang2017chestx, wang2018tienet}. In order to create diagnosis reports from images, \citet{zhang2017mdnet} propose a unified approach that employs a multimodal mapping between images and reports. They obtain an interpretable diagnosis process, by relying on attention mechanisms to highlight important aspects of the data. Similarly, Tandemnet is a neural network architecture that combines visual and textual attention to highlight regions that might be of interest to a pathologist and shows that the inclusion of textual information appears to improve visual attention results \citep{zhang2017tandemnet}. In a more general setting, \citet{zagoruyko2016paying} compare the usefulness of activation-based versus gradient-based attention map mechanisms and find that the activation-based mechanism leads to the best performances when classifying images with attention transfer techniques.

An issue with the interpretation of saliency maps is that they provide no information about why particular pixels are important for the outcome or to what particular class they belong. While the maps indeed point to important evidence in the data, the amount of justification they provide is limited. Our research, while also providing a unified approach across image and text, therefore focuses on an entirely new justification technique. As a part of our methodology relies on networks that are capable of generating data across different modalities, we first discuss existing techniques in these fields.

\subsection{Generative and Cross-Modal Architectures}\label{sec_gen}

Recent advances in generative models have mostly been driven by Generative Adversarial Networks (GANs) \citep{2014_GAN_original_paper}. With adversarial networks, a generator is trained to produce realistic outputs that mimic the distribution of the training data. This is achieved by simultaneously training a discriminator that attempts to distinguish real images from generated ones. Several formulations of discriminator training objectives exist, where in particular the Wasserstein GAN formulation is of interest \citep{2017_GAN_wasserstein} as its interpretation as a distance between distributions allows it be used as an evaluation metric \citep{danihelka2017comparison,im2018quantitatively}.

A well-known example of a model that successfully applies adversarial training for image generation is the Deep Convolutional Generative Adversarial Network (DCGAN) \citep{2015_GAN_DCGAN_ICLR} that consists of a series of convolutional layers. Other methods have built on such techniques, for example by generating images conditional on other information \citep{odena2017conditional}, or by progressively growing the images by conditioning on lower-dimensional output \citep{zhang2016stackgan, Karras_Aila_Laine_Lehtinen_2017}. These approaches have thus made detailed high-resolution category-dependent image generation possible. 

During training of conditional GANs, the class information is passed along to both generator and discriminator so that the networks implicitly learn to classify samples. This insight is also used in translation tasks from one domain to another. An example is the mapping of images to other images where only a particular aspect of the data is transformed while maintaining non-category dependent details \citep{choi2017stargan}. \citet{zhu2017unpaired} have shown that adding a cycle-consistency loss can help when translating data from one domain to another. Such a loss demands consistency when mapping from a domain $X$ to a domain $Y$ with a mapping $F(X)$ by requiring that $G(F(X)) \approx X$ when projecting back to the input domain.  

When training to classify data, a neural network internally learns to map the high-dimensional input data to a low-dimensional manifold. Ideally those manifolds should be continuous surfaces such that one smoothly and sensibly migrates from one category to another while traversing such a manifold. An interesting problem that arises with deep neural networks however is that they tend to learn mappings that are fairly discontinuous. As a result, images that contain a limited amount of noise in certain parts of the image, so-called adversarial examples, might lead a neural network to completely misclassify an image \citep{szegedy2013intriguing}. GANs can also be used as a successful strategy to avoid such outcomes. By learning to generate a realistic distribution of unperturbed images with a GAN setup, one can then restrict the solution space to the output of the generator \citep{samangouei2018defense}. 

GANs are particularly difficult to train for discrete distributions, such as textual data, but advancements have recently been made in this area as well \citep{che2017maximum}. Typically the trouble with training textual GANs is that it is hard to obtain a smooth latent surface as the differentiation operator during back-propagation over discrete symbols can lead to large gradients and thus unstable training. A solution to this problem is to employ gradient penalties to ensure stable training \citep{gulrajani2017improved}. Another possibility is to train Adversarially Regularized Autoencoders (ARAE), which are autoencoders that are regularized with an adversarial setup \citep{junbo2017adversarially}. The latter has the advantage of creating a continuous representation space for the discrete input that is smooth in the sense that sentences with similar content are located near each other in the latent space. This method has also been shown to produce high-quality textual output compared to more traditional RNN generation methods based on maximum likelihood \citep{spinks2018generating_naacl}. 

Of particular interest in this research are networks that operate across different modalities such as image and text. One interesting example is the progressively grown StackGAN-v1 \citep{zhang2016stackgan}, which requires a continuous textual representation at its input. This network then generates images that take the conditional information into account in two stages. First a low-resolution image is created that defines the main features and attributes of the output. In a second stage, a high-resolution image is created conditional on both the textual and low-resolution input. The StackGAN-v1 architecture also adds a conditioning augmentation system that maps the textual input to a multivariate Gaussian distribution from which one samples a new textual representation. The goal of this procedure is both to augment the data as well as to create smooth interpolations between different sentences. 

In the following section we will explain how we combine these techniques in a unified approach for our diagnosis and justification system.

\section{Method}\label{method}

The methodology consists of a training phase and an inference phase. During the training phase, the focus lies on optimizing the different networks of the system. First, continuous textual representations, which we refer to as embeddings, are learned with an ARAE for descriptions that consist of the concatenated diagnosis label and caption text of X-Ray images. The embeddings are concatenated to noise vectors, thus forming the representations that function as the bridge between image and text. From those representations we train a text-to-image GAN to generate realistic X-Ray images that mimic the distribution of the training set. Finally we train a convolutional network to perform the inverse mapping, $I$, that projects X-Ray images to continuous representations. 

During the inference stage, the inverse mapping $I$ is applied to a real X-Ray to obtain a continuous representation $r$ from which a diagnosis $d$ and textual justification in the form of a caption $c$ can be retrieved. Additionally, a mapping $M$ is performed that transforms the representation $r$ to a representation $r'$ under constraints that ensure that the X-Ray that can be generated from $r'$ displays an alternative diagnosis $d'$ while maintaining as much visual similarity as possible to the original (real) X-Ray. This offers a justification that visually explains why the network chose one diagnosis over its nearest contender.

\begin{figure}
	\centering
	\includegraphics[width=0.999\linewidth]{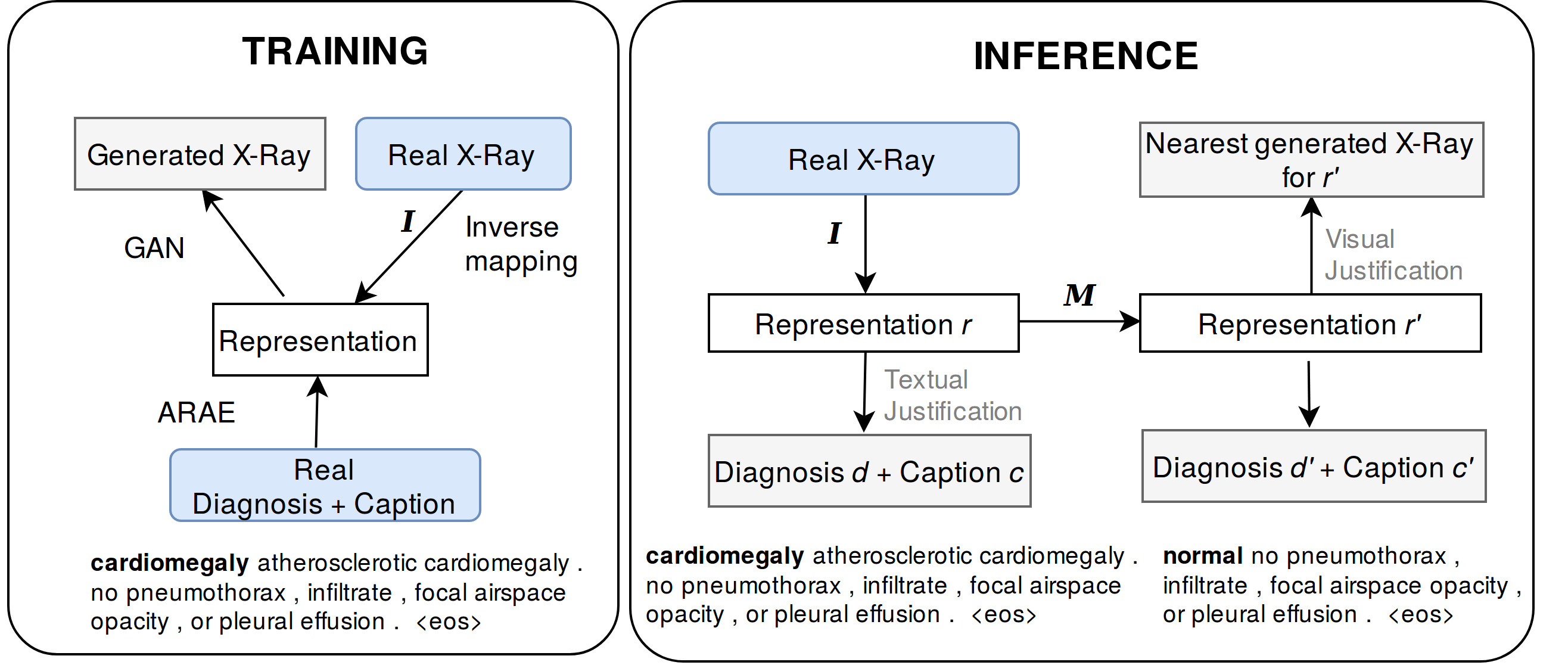}
	\caption[methodology]{Overview of the entire methodology. During the training phase, the different networks are trained in several steps: first an ARAE learns to create the smooth textual embedding from which we obtain a representation; next, a GAN learns to produce images from the representations; and finally, a convolutional neural network for the Inverse mapping ($I$) learns to obtain representations from X-Rays. During the inference phase, the Inverse mapping ($I$) provides a representation ($r$), from which we diagnose and textually justify that diagnosis ($d$) with a caption ($c$) for a real X-Ray. An additional mapping ($M$) is also performed during the inference stage by directly transforming the representation ($r$) into a different representation ($r'$) that belongs to the nearest alternative diagnosis ($d'$). From the obtained representation ($r'$), an X-Ray is generated with the trained GAN in order to provide an additional visual justification. }
	\label{fig:methodology}
\end{figure}

The entire methodology is illustrated in figure \ref{fig:methodology} and will be explained in detail in the coming sections. In table \ref{tab:notations} we provide an overview of the notations that are used in this section.

\begin{table}[t!]
	\begin{center}		
		\begin{tabular}{|c|l|}
			\hline Term & Explanation\\
			\hline
			ARAE & Adversarially Regularized Autoencoder \citep{junbo2017adversarially}\\ 
			embedding & The textual embedding that is obtained by training\\& the ARAE model \\		
			$v$ & The dimension of the embeddings \\	
			$r$ & The vector representation that is obtained by concatenating\\& the embedding with a noise vector 	\\
			$I$ & The inverse mapping, performed by a convolutional network,\\& from X-Ray to representation $r$\\
			$d$ & The diagnosis for the X-Ray, obtained by decoding $r$\\
			$c$ & The caption for the X-Ray, obtained by decoding $r$, serves\\&  as a textual justification\\
			$r'$ & The vector representation that is similar to $r$ but contains\\&  an alternative diagnosis $d'$\\
			$d'$& The nearest diagnosis to the obtained diagnosis $d$ for this\\&  X-Ray, obtained by decoding $r'$ \\
			$c'$&The caption obtained by decoding $r'$\\
			$M$&The mapping that is performed that directly modifies $r$ to\\&  obtain $r'$ under a set of constraints\\
			GAN & The text-to-image Generative Adversarial Network that is\\& used  to create images from representations such as $r$ and $r'$\\
			
			\hline
		\end{tabular}
	\end{center}
	\caption{\label{tab:notations} Overview of the notations used in section \ref{method}.  }
\end{table}

\subsection{Creating a continuous textual representation} \label{ARAE_method}

The first step in our method regards the training of an ARAE (see section \ref{sec_gen}), which consists of an autoencoder for sentences trained in an adversarial setup. As the architecture we use the same setup as in \citep{junbo2017adversarially} where both the encoder $E$ and decoder $D$ are single-layer LSTM networks and the optimizer is the Adam optimizer with standard beta parameters $0.5$ and $0.999$. For each image in the data set, descriptions are created by concatenating the diagnosis (label) to the caption, which consists of the medical findings that describe the content of the image. During training of the ARAE, continuous embeddings (vectors with a dimension $v$) of these descriptions are created at the output of the hidden layer of the encoder. The embeddings are regularized by constraining the training in an adversarial setup where a generator and a critic, that are fully-connected networks consisting of three layers, force the learned embedding to be smooth. We found that $v=300$ leads to good results in terms of discriminator convergence and autoencoder reconstruction. The model with the lowest reconstruction error on the validation set over 300 epochs is selected.

The representations that will serve as an intermediate space between sentences and images are then obtained by concatenating the embedding to a noise vector. In these representations, the embeddings embody the semantic (high-level) content while the noise vectors embody the visual (low-level) structure such that two identical sentences may still lead to very different X-Rays. In our experiments we use normally distributed noise vectors of dimension 100 (see also section \ref{representation_exp}). 

Constructing such representations is crucial to our approach. As shown by \citet{junbo2017adversarially}, the sentence embeddings learned by the ARAE model map similar sentences to a similar space in the representation space. This will allow us to create better quality images that take into account the conditional information in the textual representation. Additionally, our mapping procedure (see section \ref{mapping}) relies on this semantic similarity of nearby representations. 

In the next section, we will create images from the learned representations with a text-to-image model.

\subsection{Text-to-Image GAN}\label{GAN}
For the text-to-image model a similar architecture is used to the StackGAN-v1 which was detailed in section \ref{sec_gen}. At the input of the architecture we present the sentence representations consisting of the embeddings we learned in section \ref{ARAE_method}, concatenated to a normally distributed noise vector. The original StackGAN architecture is simplified however by removing the conditioning augmentation mechanism.  The textual representations that are created with the ARAE in section \ref{ARAE_method} are already smooth and thus there is limited benefit from such an augmentation. Additionally, we avoid the resampling mechanism which would complicate the mapping procedures that we will detail in the following sections. The details of the StackGAN architecture that we use can be found in \ref{app:stackgan}.

There are two important reasons to use a GAN as the text-to-image model. Firstly, it has been demonstrated that GANs are able to create high-quality outputs, even at high resolutions. Secondly, the ability of GANs to produce examples that don't occur in the training data is essential to the visual justification that will be explained in section \ref{mapping}.

In our setup we first train the network to generate images with relevant features at a resolution of $64\times64$ pixels conditional on the textual input. In the second stage of training, images with a resolution of $256\times256$ pixels are generated, conditioned both on the textual input and the lower-resolution images that were created in the first stage. This two-stage generation process allows for highly detailed outputs at an acceptable resolution. While the original data set contains images with resolutions of $1024\times1024$ pixels, we limit the maximum resolution to keep GPU computations feasible. The same optimizer and training regime are used as in the StackGAN-v1 paper \citep{zhang2016stackgan}.

\subsection{Inverse mapping $I$ to the continuous code space} \label{invG}

Upon training the GAN, we perform a mapping from the visual input to the continuous representation space that was created in section \ref{ARAE_method}. In our setup, this projection is the inverse mapping of the generator $G$ from section \ref{GAN}, i.e., after learning to generate images from the continuous vectors, the existing images are projected back to the lower-dimensional continuous manifold. Once this mapping is learned, we split the representation into noise vector and embedding. The embedding is then decoded with the decoder $D$ that was learned as part of the ARAE of section \ref{ARAE_method}. This method is similar to the one used by \citet{spinks2018generating} to create accompanying captions for a set of medical images. 

As the two stages in the image generation of StackGAN-v1 setup are decoupled, this inverse mapping is performed on the output of the first stage (with resolution $64\times64$) rather than the second stage (with resolution $256\times256$). Our tests show no loss in categorization result for the given labels and it has the added benefit of a much leaner and faster architecture. 

As the word that is first decoded from the continuous embedding always represents the diagnosis label, we get an implicit estimate of the certainty of the diagnosis, for example: $70\%$ cardiomegaly, $30\%$ normal. By continuing the decoding process, conditional on the label, a textual justification for the diagnosis is obtained. An example might look as follows:

cardiomegaly: heart size slightly elevated. EOS 

\subsubsection*{Mathematical outline}

The mapping to the continuous representation should not only lead to correct diagnoses but also to captions that are consistent with the evidence in the image. In order to obtain this result we train with a combination of several loss functions: 

\begin{enumerate}
	\item A label loss that minimizes the diagnosis outcome ($\mathcal{L}_{diagnosis}$).
	\item A cycle-consistency loss \citep{zhu2017unpaired} that minimizes the image reconstruction when generating an image from the representation $r$ that was obtained from an image $x$ ($\mathcal{L}_{cycle\_img}$ such that $G(I(x)) \approx x$).
	\item  A cycle-consistency loss \citep{zhu2017unpaired} that minimizes the reconstruction of the embedding obtained that is part of the representation $r$ from which an image was originally created ($\mathcal{L}_{cycle\_emb}$ such that $I(G(embedding)) \approx embedding$).
\end{enumerate}
  
The latter loss also ensures that the mapping results in a realistic embedding from the originally learned embedding space. To make sure the solution doesn't evolve towards a local optimum in a bad part of the solution space, the diagnosis loss $\mathcal{L}_{diagnosis}$ is only applied progressively with a weight of $log_{10}(epoch+1)$. We found empirically that such a logarithmic term leads to good results as only a few epochs are needed to find a good starting position in the solution space. The entire loss is thus summarized in equation \ref{eqn:inv_loss}. 
  
\begin{equation}\label{eqn:inv_loss}
\mathcal{L}_{I} = log_{10}(epoch+1)\times \mathcal{L}_{diagnosis} +  \mathcal{L}_{cycle\_img} +\mathcal{L}_{cycle\_emb}
\end{equation}

where $\mathcal{L}_{diagnosis}$ is the cross-entropy loss between label categories, $\mathcal{L}_{cycle\_img}$ is the mean square error of the reconstruction in the pixel space, i.e., it represents how different each pixel value in the reconstruction is from the original. Finally, $\mathcal{L}_{cycle\_emb}$ is the mean square error in the embedding space, which measures the distance between both vectors in the $v$-dimensional space. The architecture for the deep neural network that performs this inverse mapping relies on a series of convolutional blocks and fully-connected layers. The details can be found in \ref{app:inv}. The optimizer is the Adam optimizer with standard beta parameters $0.5$ and $0.999$. The best model is selected based on the best diagnosis accuracy on the validation set over 100 epochs.

In the next section we show that an extension to this setup also provides a new type of visual justification. 

\subsection{Additional justification with a mapping $M$} \label{mapping}

The goal now is to present an additional justification for a human operator. The idea is to answer the following question: ``Given the diagnosis for an X-Ray that is found with the methodology of section \ref{invG}, what would the X-Ray look like for the nearest alternative diagnosis". Such justification is particularly useful for an operator who hesitates between two diagnoses. Providing an image that belongs to the nearest alternative diagnosis would let the operator understand where the network exactly differentiates between both diagnoses and why it has decided to deliver one outcome over another.

As mentioned in section \ref{sec_gen}, a concern when creating similar images is the influence of adversarial noise on classification outcomes. In other words, by directly modifying an X-Ray image with diagnsos $d$ in the pixel space we might produce an X-Ray that is very similar to the original X-Ray and decodes to the nearest alternative diagnosis $d'$, yet doesn't actually portray the characteristics of that diagnosis $d'$. For this reason, we employ a technique inspired by the work of \citet{samangouei2018defense} where the solution space of the images is restricted to the outputs defined by the generator of the text-to-image network. In essence, rather than modifying the image directly, we modify the representation $r$ that is obtained by performing the inverse mapping $I$ on a real X-Ray and that decodes to $d$. With the mapping, we obtain a representation $r'$ that decodes to $d'$ and from which an X-Ray can be generated that is very similar to the original X-Ray. As the X-Ray is generated with the conditional text-to-image GAN, the image contains characteristics that belong to the diagnosis $d'$.

This mapping $M$ is performed only at inference time and is subject to three criteria in total. The criteria are that (1) $r'$ should be similar to $r$ yet (2) decode to an alternative diagnosis $d'$ while (3) maintaining visual similarity in the associated X-Ray that is generated from it. The mathematical details are given below.

\subsubsection*{Mathematical outline}

Name $r$ the representation that was found in section \ref{invG} from a real X-Ray $x$ that contains the textual embedding that decodes to a diagnosis or label $d$. Now we create a continuous vector $r'$ that we initialize with the value of $r$ and then modify with $L$ steps of a simple gradient descent until we find an optimum under the following constraints. First, the embedding in the obtained representation should decode to a diagnosis $d'$ from the nearest alternative class. The loss is thus given by the cross-entropy over the conditional probability that is defined by the decoder as shown in equation \ref{eqn:alt_diag}. Second, the X-Ray $x'$ that is generated from this representation $r'$ with the generator $G$ of the text-to-image model should be similar to the original X-Ray $x$. This can be implemented with a mean square error in pixel space as shown in equation \ref{eqn:img_sim}. Finally, the modified representation $r'$ should be relatively close to the original representation $r$ in order to avoid degenerate solutions. This is also implemented with a mean square error (equation \ref{eqn:emb_sim}). If the initialization of $r'$ from $r$ doesn't reach an optimum we try again by initializing from $r+z$ where $z$ is a small random normal vector. The entire loss is thus given by equation \ref{eqn:M_loss}. 

\begin{equation}\label{eqn:alt_diag}
\mathcal{L}_{alt\_diagnosis} = -log(p(d')|r')
\end{equation}
\begin{equation}\label{eqn:img_sim}
\mathcal{L}_{img\_similarity} = \lVert G(r') - x \rVert^2_2
\end{equation}
\begin{equation}\label{eqn:emb_sim}
\mathcal{L}_{emb\_similarity} = \lVert r' - r \rVert^2_2
\end{equation}
\begin{equation}\label{eqn:M_loss}
\mathcal{L}_{M} = \mathcal{L}_{alt\_diagnosis} + \mathcal{L}_{img\_similarity} + \mathcal{L}_{emb\_similarity}
\end{equation}

To find the optimum under these loss functions, a L-BFGS optimizer is used with learning rate 1 over 100 gradient descent steps with maximum 20 iterations per step.

We have now detailed the entire methodology for our unified approach. It consists of the construction of a smooth representation space of the text that acts as an intermediate space between text and image (section \ref{ARAE_method}), the training of a text-to-image model (section \ref{GAN}) and the training of the inverse mapping $I$ back to the continuous space (section \ref{invG}). At inference time, $I$ is performed on a real X-Ray thus delivering a representation from which the diagnosis label and textual justification can be found. Additionally, the mapping $M$ is performed that provides an additional visual justification (section \ref{mapping}).

\section{Experiments}

The main focus of our evaluation is on the added value of the justification mechanism in section \ref{justification_exp}. While our method combines several existing methodologies in a unified approach, we also shortly evaluate the performance of the individual parts, i.e., the learning of a continuous representation (\ref{representation_exp}), diagnosis from image (\ref{diagnosis_exp}) and image captioning (\ref{captioning_exp}). The analysis is performed on a set of frontal chest X-Rays from the data set of the National Library Medicine, National Institutes of Health, Bethesda, MD, USA \citep{demner2015preparing}.

\subsection{Preprocessing} \label{preprocessing}

For each X-Ray, a label for a diagnosis and the corresponding findings that describe visual evidence for the diagnosis are concatenated into a textual description. As the sentences that describe the visual evidence are almost entirely complementary and disjoint, we create up to four distinct captions for each X-Ray with a maximum length of 15 words each. This is done by splitting the original description into sentences that are recombined into distinct captions. Captions that contain less than 5 words are discarded and words with less than 20 occurrences are replaced by out-of-vocabulary markers. The images are scaled to the desired size and slightly cropped in order to augment the visual data for training. We divide the data into random training, validation and test sets according to the following weights: $80\%$, $10\%$ and $10\%$ \footnote{We will make these splits available upon acceptance.}. If one of these subsets (for example the training set) contains an X-Ray for a particular patient, the other subsets (validation and test set) will not contain any X-Rays for that same patient.

As mentioned before, one of the difficulties with medical image diagnoses is that one image might contain several labels for distinct disease patterns. Additionally, the annotations for some of these data sets are questionable. In order to clearly illustrate the validity of our justification technique we limit the data set to images that contain no indications of diseases (labeled as ``normal") and images that exhibit cardiomegaly (labeled as ``cardiomegaly"). The latter is an affliction characterized by a disproportionally enlarged heart which should remain discernible at the resolutions that are used in this work. The labels are highly unbalanced with roughly $84\%$ of the data containing the label ``normal" and $16\%$ containing the label ``cardiomegaly". Note that while we illustrate the methodology in this paper on these two labels, this method can be applied to data sets with any amount of labels. After applying the preprocessing, the resulting data set contains 2446 X-Rays with 4725 textual descriptions.

\subsection{Representation learning} \label{representation_exp}

As explained in section \ref{ARAE_method}, the first step in our approach involves creating a continuous and smooth textual embedding with the ARAE model. To evaluate the benefit of such a representation space, we compare it to a setup where, during the image generation step, the textual embedding is learned in an embedding layer directly from one-hot encoded word inputs. We then compare the outputs of the text-to-image GAN for both methods.

To evaluate the quality of the outputs, we use the Wasserstein distance estimation as an evaluation measure as proposed by \citet{danihelka2017comparison}. The Wasserstein distance is an approximation of the Earth Mover Distance which quantifies the amount of effort that is needed to move the mass of one distribution to another. Thus if the quality of the outputs in the distribution of generated X-Rays approaches the quality of the X-Rays in the test set, the Wasserstein distance will approach zero. The Wasserstein distance is obtained by training a discriminator with the Wasserstein loss (see section \ref{sec_gen}) on the validation set and subsequently evaluating it on the test set \citep{danihelka2017comparison}. Additionally we compute the 'alignment' formulation of this distance that takes into account the conditional information and thus gives an estimate of how well the conditional information was assimilated in the image generation process \citep{spinks2018evaluating}. In table \ref{res:representation}, lower numbers represent better convergence and thus better outcomes. It becomes apparent that the ARAE embeddings lead to vastly better image quality in the stage-I generated images compared to a simple embedding layer. These results underline the necessity of creating a smooth representation space for natural language and the ARAE embeddings are therefore used in the remainder of our experiments.

\begin{table}[t!]
	\begin{center}
		\begin{tabular}{|l|ccc|}
			\hline Metric & Embedding layer & ARAE & p-value 	\\
			\hline
			Wasserstein estimate & 0.38 & \bf0.19   & $<10^{-5}$ \\
			Alignment estimate & 1.19 & \bf0.46  	& $<10^{-5}$ 	\\
			
			\hline
		\end{tabular}
	\end{center}
	\caption{\label{res:representation} Estimates of the Wasserstein distance and alignment over 4 runs for images generated with an embedding layer versus images created with learned ARAE embeddings. The outcomes are estimations of the Wasserstein distance, where smaller numbers are better. The corresponding p-value is also given for a two-tailed equal variance t-test.  }
\end{table}

Using the same principle, we investigate the effect of concatenating different noise vectors to the embeddings. We modify the dimension of the noise vector in increments of 50 between 50 and 300 and concatenate them to the learned embeddings. As noise vectors, we both try normally distributed and uniformly distributed vectors. We then again evaluate the Wasserstein and alignment estimate on the created images. The results show that quality and alignment deteriorate for dimensions larger than 200. Additionally, uniform noise leads to similar output quality and alignment as compared to normally distributed noise. A dimension of 100 leads to good results both for normally and uniformly distributed noise. For the remainder of the experiments we use normally distributed noise with a dimension of 100.

\subsection{Diagnosis} \label{diagnosis_exp}

\begin{table}[t!]
	\begin{center}
		\begin{tabular}{|l|cc|}
			\hline Model & Accuracy & $\sigma$  \\
			\hline
			Vgg19 & 0.843 & 0.054   \\
			Resnet34  & 0.90 & 0.016  \\
			Inception & 0.907& 0.015  		\\
			Alexnet&0.910& 0.008  		\\
			Dense-net&0.914& 0.015  		\\
			Resnet18 & \bf0.919& 0.007  		\\
			Ours & 0.906& 0.008  		\\
			
			\hline
		\end{tabular}
	\end{center}
	\caption{\label{res:diagnosis} Classification accuracy for different classification models.   }
\end{table}

Given the concerns over the quality of the data set labels, the absolute level of the diagnosis accuracy is not the main concern in this work, but considering the multiple components of our setup and the multi-task training, it is important to verify that the diagnosis accuracy doesn't deteriorate compared to single-task setups.

\begin{figure}
	\centering
	\begin{subfigure}{.5\textwidth}
		\centering
		\includegraphics[width=1.0\linewidth]{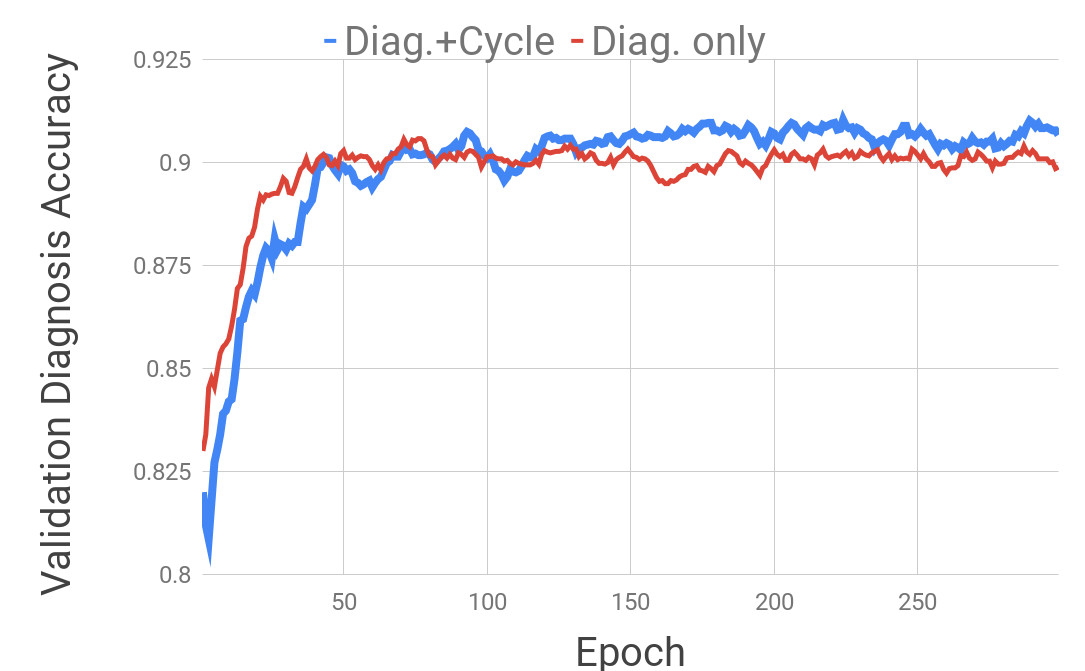}
		\caption{Diagnosis accuracy for all epochs.}
		\label{fig:sub1}
	\end{subfigure}%
	\begin{subfigure}{.5\textwidth}
		\centering
		\includegraphics[width=1.0\linewidth]{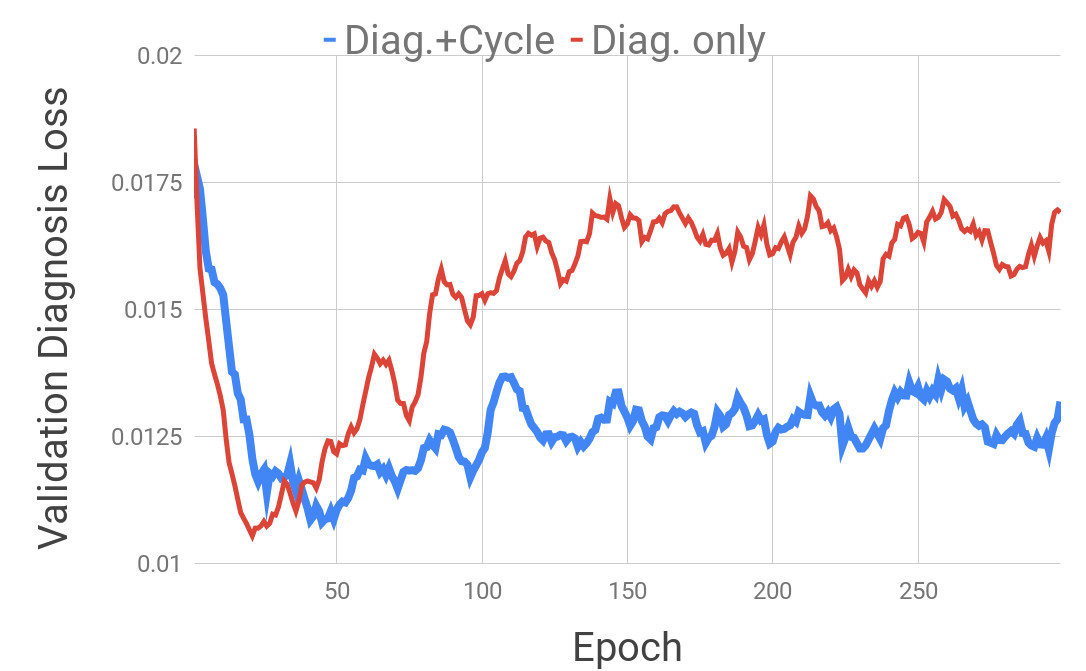}
		\caption{Diagnosis loss for all epochs.}
		\label{fig:sub2}
	\end{subfigure}
	\caption{When tracking the accuracy and loss of the diagnosis of the images in the validation set, we notice that including cycle-consistency losses during training, rather than only using the label loss, does not deteriorate performance. (Graphs are smoothed by averaging over the ten nearest datapoints.)}
	\label{fig:inv_overfitting}
\end{figure}

In figure \ref{fig:inv_overfitting}, it is shown that the multi-task training that consists of both classification and cycle-consistency training, during the inverse mapping $I$ doesn't deteriorate diagnosis accuracy and even stabilizes the diagnosis loss to a certain extent over longer training periods. 

Additionally, we compare the classification accuracy on the test set of our multi-task inverse mapping $I$, to several recent state-of-the-art classification methods that are trained using only the classification loss. We see in table \ref{res:diagnosis} that our model holds up well among these baselines. Note that the Vgg19 model training failed to obtain good results in several runs, leading to an overall lower accuracy and higher standard deviation.

The suggested method thus achieves a mapping to the representation space with similar diagnosis accuracy, roughly $91\%$ on the test set, with the additional benefit of mapping to a learned representation space. Given the cycle-consistency loss training, the latter allows us to decode the embeddings in those continuous representations into captions containing diagnosis information and textual evidence.

\subsection{Captioning} \label{captioning_exp}
\begin{table}[t!]
	\begin{center}
		\begin{tabular}{|l|cccccccc|}
			\hline Model & B1 & B2 & B3 & B4 & R& M&C &$\sigma$\\
			\hline
			\citep{aneja2018convolutional} w. pretr. & 0.34 & 0.23   & 0.16  & 0.12 & 0.29   &0.15&0.54&0.03\\
			\citep{aneja2018convolutional} w/o pretr.  & 0.39 & 0.26   & 0.20  & 0.15 & 0.32   &0.15&\bf0.64&0.00\\
			Ours-label & \bf0.49& \bf0.35  	& 0.24  & \bf0.18 & \bf0.40   & \bf0.27&0.58&0.02	\\
			Ours-caption & \bf0.49& \bf0.35  	& \bf0.25  & \bf0.18 & \bf0.40   & \bf0.27&0.60	&0.01\\
			
			\hline
		\end{tabular}
	\end{center}
	\caption{\label{res:captioning} Results for image captioning experiments where B1 = BLEU1, B2 = BLEU2, B3 = BLEU3, B4 = BLEU4, R = Rouge, M = Meteor, C = CIDEr, $\sigma$ = standard deviation for the CIDEr metric over three runs. The results of the convolutional captioning architecture of \citep{aneja2018convolutional} with and without pretraining are compared to our method where the best model is selected based on either the label accuracy or the caption score on the validation set. }
\end{table}

As explained in section \ref{invG}, our method obtains captions after performing the inverse mapping $I$. 
To verify that these captions are of good quality, we compare them to captions that are obtained by a general baseline that is trained to produce captions directly from the X-Rays. We use a convolutional captioning model with attention mechanism that was recently shown to outperform the traditional captioning setup that combines a convolutional image feature network and LSTM captioning model \citep{aneja2018convolutional}. While this model uses a pretrained ImageNet classifier network that is subsequently finetuned during training, we also compare it to a setup where the classifier network is trained from scratch as the ImageNet features would not necessarily be useful for X-Ray images. In table \ref{res:captioning} we show the outcomes for conventional metrics: the BLEU-1 through BLEU-4 metrics \citep{papineni2002bleu} as well as the ROUGE \citep{lin2004rouge}, METEOR \citep{denkowski2014meteor} and CIDEr \citep{vedantam2015cider} scores. While the BLEU and ROUGE scores focus on co-occurrence statistics based on n-grams, METEOR aligns the captions to reference sentences and calculates sentence-level similarity scores. CIDEr is an evaluation score that is specifically aimed at image captioning tasks with multiple descriptive sentences per image.

We show the results for our method when the best model is selected based on the best validation label accuracy (``Ours-label") and when it is selection based on the best validation CIDEr score (``Ours-caption") as was the case for the baseline models. Both our models perform similarly and achieve a good CIDEr score. Our models outperform the baselines on all other metrics. 

\begin{figure}
	\centering
	\includegraphics[width=1.0\linewidth]{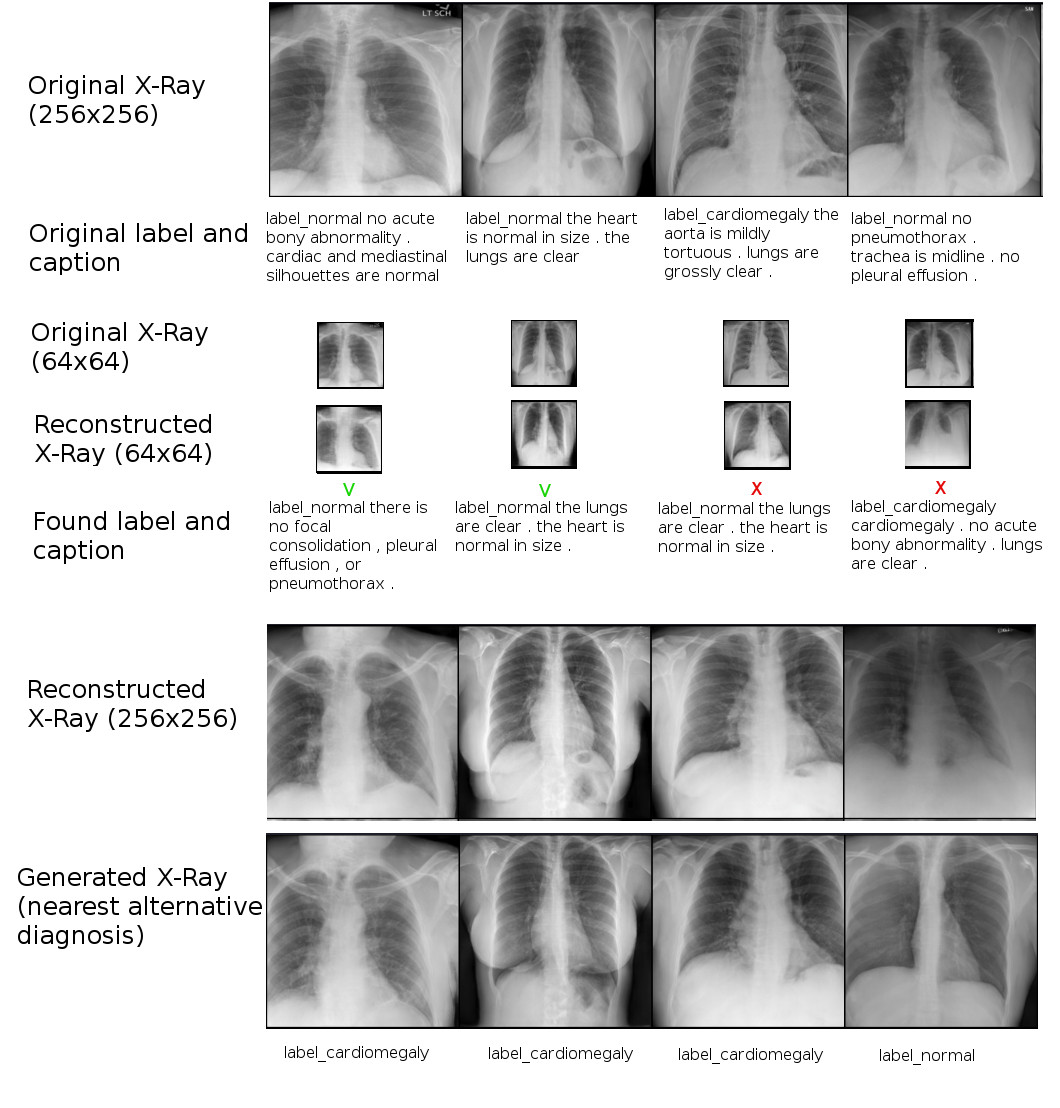}
	\caption[outcomes]{Examples of outcomes for a set of X-Rays. Under the original image and caption, the reconstruction of the original X-Rays at a resolution of $64\times64$ is shown as well as the obtained diagnosis and caption. For the first two X-Rays, the outcome is the same as the ground truth labels while the last two are examples where a different diagnosis is found. In the bottom 2 rows, the reconstructed X-Ray at a resolution of $256\times256$ is shown together with the generated X-Ray for the nearest alternative diagnosis.}
	\label{fig:outcomes}
\end{figure}

\begin{figure}
	\centering
	\includegraphics[width=1.0\linewidth]{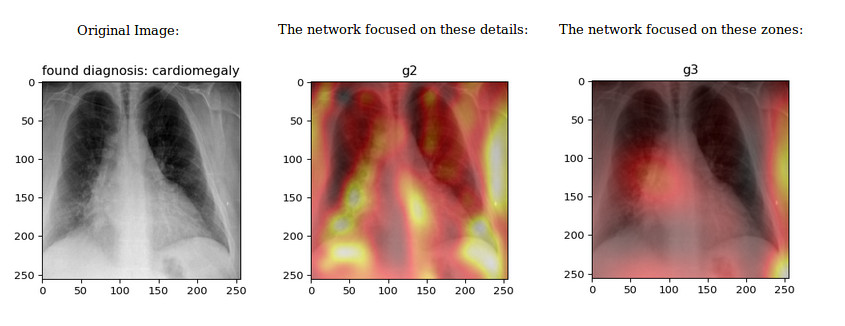}
	\caption[alternative_method]{Example of the alternative diagnosis method that relies on heatmaps to demonstrate where the neural network focuses its attention to reach a diagnosis. Saliency maps are provided both at a more detailed level as well as at a more abstract level. }
	\label{fig:alt_method_example}
\end{figure}

\subsection{Justification} \label{justification_exp}

Applying the inverse mapping $I$ on real X-Rays from the test set provides us with representations from which a diagnosis and textual justification is obtained. Additionally, we perform the mapping $M$ on the test set to provide an additional visual justification. Note that while $M$ mostly needs just one try in order to converge to a solution under the constraints detailed in section \ref{mapping}, some batches need more attempts. All batches in the test set eventually reach a stable solution in less than 20 tries. Some examples of different samples of the test set are shown in figure \ref{fig:outcomes}.

We also compare our method to the alternative approach that relies on visual heatmaps. For this alternative, we train a ResNet of 18 layers deep \citep{he2016deep} that obtains the best accuracy for this data set in our experiments in section \ref{diagnosis_exp}. We extract the saliency maps from the activation-based attention schemes, as suggested by \citet{zagoruyko2016paying} at both an intermediate level of the ResNet architecture as well as at a more abstract level. This method thus provides visual evidence for a human operator of what elements of the image are most important for the proposed diagnosis. An example is shown in figure \ref{fig:alt_method_example}.

The goal of this evaluation is now to determine the value of the proposed alternative justification mechanisms in a realistic diagnosis situation with a human expert operator. We serve three trained clinicians some questions that attempt to determine the usefulness as well as the validity of the given justification. In total 300 thorax X-rays with a resolution of $256\times256$ pixels are analyzed from the test set, 150 of which are handled by the 'saliency map' method while the remaining 150 are handled by our method. 50 of the X-Rays for each method are then examined by each expert and for 30 X-Rays we asked the opinion of all three. The following questions are asked for each method:

\begin{enumerate}	
	\item Does the presented evidence help explain why this diagnosis was put forward? 
	(on a scale of 1 to 4  where 1= No, not at all; 4 = Yes, very much)
	\item Does the network appear to have a correct understanding of which parts of the X-Ray are important for the diagnosis? 
	(on a scale of 1 to 4  where 1= No, not at all; 4 = Yes, very much)
	\item Do you agree with the proposed diagnosis?
	(0 = No, 1 = Yes)
	\item How certain are you about your final diagnosis?
	(on a scale of 1 to 4 where 1= not sure at all; 4 = very sure)
\end{enumerate}

Question 1 explicitly attempts to determine to what degree the neural network explains itself and thus is the main metric we are concerned with. Question 2 examines whether the approach has learned the correct indicators for a diagnosis based on the provided evidence. As both the network that uses the saliency map method as well as the one that uses our justification method  achieve similar accuracy scores in terms of diagnosis, we expect a priori that the outcome for questions 2, 3 and 4 will be similar for both methods. That being said, a low score on question 2 might also be the result of a low score on question number 1. Additionally, a network that generalizes better based on symptom markers might perform better on questions 2 and 3 given the quality issues of the data set. A better justification mechanism might also be able to slightly improve the confidence of the human expert when answering question 4. 
\begin{table}[t!]
	\begin{center}
		\begin{tabular}{|l|ccc|}
			\hline Question & Saliency map & Our & p-value\\ 
			&	method		& method    &	\\
			\hline
			Q1 (Justification, scale 1-4) & 1.31 & \bf2.39 & $<10^{-10}$  \\
			Q2 (Understanding, scale 1-4) & 1.81 & \bf2.45 & $<10^{-10}$   \\		
			Q3 (Agreement, scale 0-1) & \bf0.89& 0.88 & 0.86  \\
			Q4 (Human certainty, scale 1-4) & 3.75 & \bf3.75 & 0.91  \\
			\hline
		\end{tabular}
	\end{center}
	\caption{\label{res:just} Average score given by human experts for each method for 4 different questions. For all questions, a higher score is better. The corresponding p-value is also given for a two-tailed equal variance t-test. }
\end{table}

The accumulated average scores of all clinicians are shown in table \ref{res:just}. All p-values are calculated with two-tailed equal variance t-tests. Note that the proposed justification method significantly outperforms in the task of explaining the diagnosis that was put forth as measured by question 1. When analyzing the answers of each clinician separately, the results hold. For the 30 X-Rays that are analyzed by all three experts, we calculate the inter-annotator agreement in the form of Fleiss' kappa for the first three questions. The outcomes are 0.33, 0.42 and 0.55 respectively, demonstrating that the evaluation is balanced and consistent across all human experts. 

The obtained results thus suggest our approach is a compelling improvement over existing approaches in justifying classification outcomes and could help human operators interpret neural network outcomes for critical tasks. Our method also appears to have a significantly better understanding of the underlying structure of the diagnosis, as measured by question 2. We suspect however that, as both methods achieve similar classification scores, the improvement in question 2 is at least partly due to the superior justification of our method as measured by question 1. In terms of how much the clinicians agree with the obtained diagnosis, as well as how certain they are about their own final diagnosis, measured by questions 3 and 4 respectively, the results are not significantly different. As the performance on the test set of both diagnosis methods is similar (see section \ref{diagnosis_exp}), we can indeed expect that the agreement with the human experts would be roughly the same in question 3. Regarding question 4, human certainty levels are relatively high across the board and thus a significant difference is less likely to be found. Overall, the improved outcomes in questions 1 and 2 underline the validity and usefulness of our approach.

\section{Conclusions}

While neural networks for medical diagnosis have become exceedingly accurate in many areas, their ability to explain how they achieve their outcome remains problematic. In this article we propose a novel method to justify the diagnosis of a neural network for medical images.  While current justification techniques highlight parts of the input that are important for the diagnosis, our method combines a textual justification with a technique that shows a visual output for the nearest alternative diagnosis according to the network.  In terms of neural network justification, our method significantly outperforms a method based on traditional saliency maps as tested in a human expert opinion study. Additionally, our approach performs very well on individual diagnosis and image captioning outcomes when compared to recent single-task baseline results. Our method is demonstrated on a data set of thorax X-Rays but can be applied to any diagnosis setup with visual input as well as classification on data sets outside the field of medicine. Better justification techniques as the one presented in this article might help facilitate the wider use of machine learning algorithms in critical medical applications.

\section*{Acknowledgments}
This  work  is part of the KU Leuven IWT-SBO 150056 project ``ACquiring  CrUcial  Medical  information Using LAnguage TEchnology” (ACCUMULATE) as well as the FWO-SNSF G078618N project on ``Deep Learning for Generating Template Pictorial and Textual Representations".

\bibliography{mybibfile}
\newpage
\appendix 

\section{Text-to-image GAN Architecture} \label{app:stackgan}
\begin{figure}[!h]
	\centering
	\includegraphics[width=0.999\linewidth]{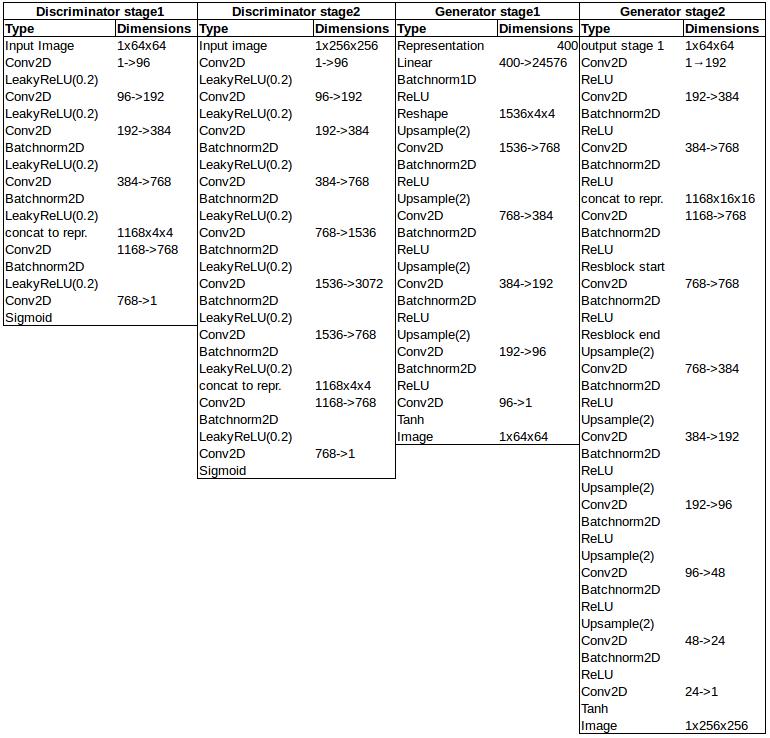}
	\caption[architecture_stackgan]{Overview of the architectures for the generators and discriminators used in the text-to-image StackGAN. Convolutional filters had a kernel size of 4 and a stride of 2. }
	\label{fig:stackgan}
\end{figure}

Figure \ref{fig:stackgan} provides an overview of the architecture that was used to train the text-to-image GAN.

\section{Inverse Mapping Architecture} \label{app:inv}
\begin{figure}[!h]
	\centering
	\includegraphics[width=0.999\linewidth]{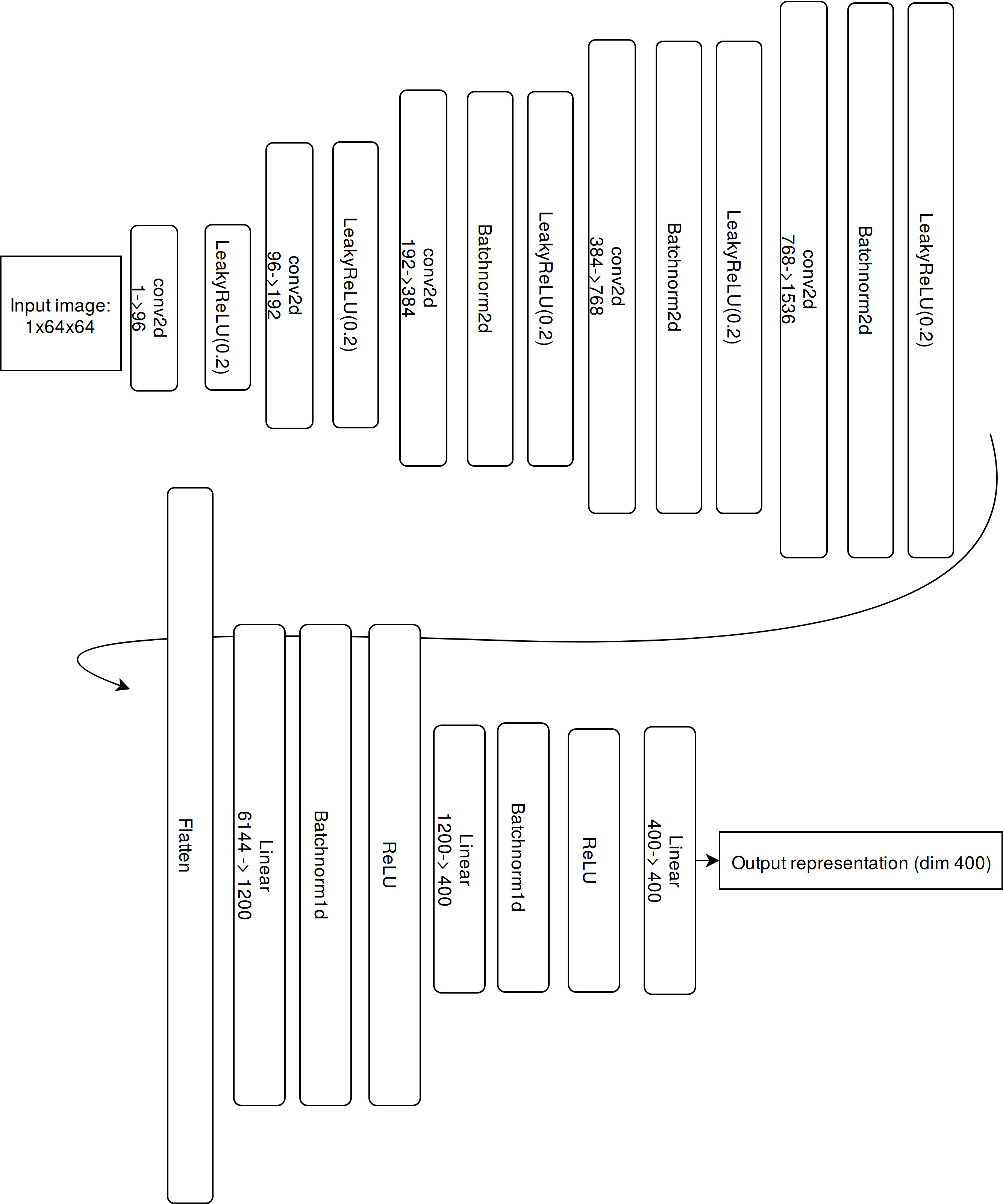}
	\caption[architecture_inv]{Overview of the architecture for the neural network used in the inverse mapping $I$. Dropout was applied after all LeakyReLU units with a probability of 0.1, and after all ReLU units with probability 0.5. Convolutional filters had a kernel size of 4 and a stride of 2. }
	\label{fig:inv}
\end{figure}

Figure \ref{fig:inv} provides an overview of the architecture that was used to perform the inverse mapping $I$.
\end{document}